\documentclass{ecai}
\usepackage{times}
\usepackage{graphicx}
\usepackage{amssymb}
\usepackage{amsmath}
\usepackage{xspace}
\usepackage{latexsym}
\usepackage[raster]{tcolorbox}
\usepackage{lstsemantic}
\usepackage{lstlangmizar}
\usepackage{booktabs}
\usepackage{url}
\usepackage{hyperref}

\def\systemname#1{\textsf{#1}\xspace}

\newcommand{\rlc}{\systemname{rlCoP}}

\newcommand{\lc}{\systemname{leanCoP}}

\newcommand{\R}{\mathbb{R}}
\newcommand{\red}{\mathop{\mathrm{red}}\limits}
\newcommand{\redd}{\mathop{\mathrm{red'}}\limits}
\DeclareMathOperator{\ReLU}{ReLU}
\DeclareMathOperator{\avg}{avg}

\definecolor{todocolor}{rgb}{1.0,0.0,0.0}

\begin{document}

\title{Property Invariant Embedding for Automated Reasoning}

\author{
  Miroslav Ol\v s\'ak
  \institute{University of Innsbruck, Austria, email: mirek@olsak.net}
\and
  Cezary Kaliszyk
  \institute{University of Innsbruck, Austria, email: Cezary.Kaliszyk@uibk.ac.at}
\and
  Josef Urban
  \institute{Czech Technical Univ. in Prague, Czechia, email: Josef.Urban@cvut.cz}
}

\maketitle
\bibliographystyle{ecai}

\begin{abstract}
Automated reasoning and theorem proving have recently become major challenges
for machine learning. In other domains, representations that are able
to abstract over unimportant transformations, such as abstraction over
translations and rotations in vision, are becoming more common.
Standard methods of embedding mathematical formulas for learning theorem proving
are however yet unable to handle many important transformations. In particular, embedding previously unseen
labels, that often arise in definitional encodings and in Skolemization, has been very weak so far. Similar problems appear when transferring knowledge between known symbols.

We propose a novel encoding of formulas that extends existing graph neural
network models. This encoding represents symbols  only by nodes
in the graph, without giving the network any knowledge of the original
labels. We provide additional links between such nodes that allow the
network to recover the meaning and therefore correctly embed such nodes
irrespective of the given labels. We test the proposed encoding in an
automated theorem prover
based on the tableaux connection calculus,
and show that it improves on the best characterizations used
so far.  The encoding is further evaluated on the premise selection task and a newly introduced symbol guessing task,
and shown to correctly predict 65\% of the symbol names.
\end{abstract}

%

\section{Introduction}

Automated Theorem Provers (ATPs)~\cite{DBLP:books/el/RobinsonV01} can be in principle
used to attempt the proof of any provable mathematical conjecture. 
The standard ATP approaches have
so far relied primarily on fast implementation of manually designed search procedures and heuristics.
However, using machine learning for guidance in the vast action spaces of the ATP calculi is a
natural choice that has been recently shown to significantly improve over the 
unguided systems~\cite{KaliszykUMO18,JakubuvU19}.

The common procedure of a first-order ATP system -- saturation-style
or tableaux -- is the following. The ATP starts with a set of first
order axioms and a conjecture. The conjecture is negated and the
formulas are Skolemized and clausified.
The objective is then to derive a contradiction from the set of clauses,
typically using some form of resolution and related inference rules.
The Skolemization as well as introduction of new definitions during
the clausification results in the introduction of many new function and predicate
symbols.

When guiding the proving process by statistical machine learning, 
the state of the prover and the
formulas, literals, and clauses, are typically encoded  to vectors of
real numbers. This has been so far mostly done with hand-crafted features resulting
in large sparse vectors~\cite{DBLP:conf/ijcai/KaliszykUV15,hammers4qed,abs-1108-3446,UrbanVS11,KaliszykU15,JakubuvU17a}, possibly reducing their dimension afterwards~\cite{ChvalovskyJ0U19}. 
Several experiments with neural
networks have been made recently, in particular based on 1D convolutions,
RNNs \cite{GollerK96}, TreeRNNs~\cite{ChvalovskyJ0U19}, and GraphNNs \cite{DuvenaudMABHAA15}. Most of the approaches, however, cannot
capture well the idea of a variable occurring multiple times in the
formula and to abstract from the names of the variables.
These issues were first 
addressed in FormulaNet~\cite{DBLP:conf/nips/WangTWD17} but even that
architecture relies on knowing the names of function and predicate symbols. This makes it 
unsuitable for handling the large number of problem-specific function and predicate
symbols introduced during the clausification.\footnote{The ratio of such symbols in real-world clausal datasets is around 40\%, see Section~\ref{symbol-experiments}.}
The same holds for large datasets of ATP problems
where symbol names are not used consistently, such as the TPTP library~\cite{Sutcliffe10}.

In this paper, we make further steps towards the abstraction of
mathematical clauses, formulas and proof states. We present a network that is invariant not only
under renaming of variables, but also under renaming of arbitrary function and predicate
symbols. It is also invariant under replacement of the symbols by their negated versions.
This is achieved by a novel conversion of the input formulas into a hypergraph, followed by a 
particularly designed 
graph neural network (GNN) capable of maintaining the invariance under negation. 
We experimentally demonstrate in three case studies
that the network works well on data 
coming from automated theorem proving tasks.

The paper is structured as follows. We first formally describe our network architecture in
Section~\ref{architecture}, and discuss its invariance properties
in Section~\ref{invariance}. We describe an experiment using the
network for guiding \lc in Section~\ref{leancop-example}, and
two experiments done on a fixed dataset in
Section~\ref{deepmath-experiments}. Section~\ref{results} contains the
results of these three experiments.

\section{Network Architecture for Invariant Embedding}
\label{architecture}

This section describes the design and details of the proposed neural architecture for invariant embeddings.
 The architecture gets as its input a set of clauses $\mathbf C$. 
It outputs an embedding for each of the clauses in $\mathbf C$, each literal and subterm and each function and predicate
symbol present in $\mathbf C$. The process consists of initially
constructing a hypergraph out of the given set of clauses, and then
several message passing layers on the hypergraph. In Section~\ref{hypergraph} we first explain the construction of a hypergraph from the input clauses. The details of the message passing are explained in Section~\ref{messagepassing}
.

\subsection{Hypergraph Construction}
\label{hypergraph}
When converting the clauses to the graph, we aim to capture as much relevant structure as possible. We roughly convert the tree structure of the terms to a circuit by sharing variables, constants and also bigger terms. The graph will be also interconnected through special nodes representing function symbols.
Let $n_{\mathbf c}$ denote the number of clauses, and let the clauses be
$\mathbf C = \{C_1, \ldots, C_{n_{\mathbf c}}\}$. Similarly, let
$\mathbf S = \{S_1, \ldots, S_{n_{\mathbf s}}\}$ denote all the function and
predicate symbols occurring at least once in the given set of clauses,
and $\mathbf T = \{T_1, \ldots, T_{n_{\mathbf t}}\}$ denote all the
subterms and literals occurring at least once in the given set of
clauses. Two subterms are considered to be identical (and therefore
represented by a single node) if they are constructed the same way
using the same functions and variables. If $T_i$ is a negative
literal, the unnegated form of $T_i$ is not automatically added to
$\mathbf T$ but all its subterms are.

The sets $\mathbf C, \mathbf S, \mathbf T$ represent the nodes of our
hypergraph. The hypergraph
will also contain two sets of edges: Binary edges
$E_{\mathbf{ct}} \subset \mathbf C\times\mathbf T$ between clauses and
literals, and 4-ary oriented labeled edges 
$E_{\mathbf st} \subset \mathbf S\times\mathbf T\times(\mathbf T\cup\{T_0\})^2\times\{1,-1\}$. 
Here $T_0$ is a specially created and added term node disjoint from all actual terms and serving in the arity-related encodings described below. The label is present at the last position of the 5-tuple. 
The set $E_{\mathbf{ct}}$ contains all
the pairs $(C_i, T_j)$ where $T_j$ is a literal contained in $C_i$. Note that this encoding makes the order of the literals in the clauses irrelevant, which corresponds to the desired semantic behavior.

The set $E_{\mathbf{st}}$ is constructed by the following
procedure applied to every
literal or subterm $T_i$ that is not a variable. If $T_i$ is a
negative literal, we set $\sigma = 1$, and interpret $T_i$ as
$T_i = \neg S_j(t_1,\ldots,t_n)$, otherwise we set $\sigma = -1$
and interpret $T_i$ as $T_i = S_j(t_1,\ldots,t_n)$, where $S_j\in\mathbf S$,  $n$ is the arity of $S_j$ and
$t_1,\ldots,t_n\in\mathbf T$. If $n=0$, we add 
$(S_j,T_i,T_{0},T_{0},\sigma)$ to $E_{\mathbf{st}}$. If
$n=1$, we add $(S_j,T_i,t_1,T_{0},\sigma)$ to $E_{\mathbf{st}}$.
And finally, if $n\geq 2$, we extend $E_{\mathbf{st}}$ by
all the tuples $(S_j,T_i,t_k,t_{k+1},\sigma)$ for $k=1,\ldots,n-1$.

This encoding is used instead of just $(S_j,T_i,t_k,\sigma)$ 
to (reasonably) maintain the order of function and predicate arguments.
For example, for two non-isomorphic (i.e., differently encoded) terms $t_1$ and $t_2$, $t_1<t_2$ will be encoded differently than $t_2<t_1$. 
Note that even this encoding does not capture the complete information about the argument order.
For example, the term $f(t_1,t_2,t_1)$ would be encoded the same way as $f(t_2,t_1,t_2)$. 
We consider such information loss acceptable.
Further note that the sets $E_{\mathbf{ct}}$, $E_{\mathbf{st}}$, and the derived sets labeled $F$ (explained below) are in fact multisets in our implementation. We present them using the set notations here for readability. 

\subsection{Message Passing}
\label{messagepassing}
Based on the hyperparameters $L$ (number of layers), and $d_{\mathbf c}^i$,
$d_{\mathbf s}^i$, $d_{\mathbf t}^i$ for $i=0,\ldots,L$ (dimensions of
vectors), we construct vectors
$c_{i,j}\in \R^{d^i_{\mathbf c}}$, $s_{i,j}\in \R^{d^i_{\mathbf s}}$,
and $t_{i,j}\in \R^{d^i_{\mathbf t}}$. First we initialize
$c_{0,j}$, $s_{0,j}$ and $t_{0,j}$ by learned constant vectors for every type
of clause, symbol, or term. By a ``type'' we mean 
an attribute based on the underlying task, see
Section~\ref{leancop-example} for an example. To preserve invariance
under negation (see Section~\ref{invariance}), we initialize all
predicate symbols to the zero vector.

After the initialization, we 
propagate through
$L$ message-passing layers. The
$i$-th layer will output vectors $c_{i,j}$, $s_{i,j}$ and $t_{i,j}$.
The values in the last layer, that is $c_{L,j}$, $s_{L,j}$ and
$t_{L,j}$, are considered to be the output of the network.
The basic idea of the message passing layer is to propagate information from a node to all its neighbors related by $E_{\mathbf{ct}}$ and $E_{\mathbf{st}}$ while recognizing the ``direction'' in which the information came. 
After this, we 
reduce the incoming data to a finite dimension using a 
reduction function (defined below)
and 
propagate through standard
neural layers.\footnote{Mostly implemented using the ReLU activation function.} 
The symbol nodes $s_{i,j}$ need particular care, because they can represent two predicate symbols at once: if $s_{i,j}$ represents a predicate symbol $P$, then $-s_{i,j}$ represents the predicate symbol $\neg P$. To preserve the polarity invariance, the symbol nodes 
are
treated slightly differently.

In the following we first provide the formulas describing the computation. The
symbols used in them are explained afterwards.

\begin{align*}
c_{i+1,j} &= \ReLU(
  B^i_{\mathbf c} + M^i_{\mathbf c}\cdot c_{i,j} +
  M^i_{\mathbf{ct}}\cdot \red_{a\in F^j_{\mathbf{ct}}}(t_{i,a})
)\cr
x_{i}^{a,b,c} &=
  B^i_{\mathbf{ts}} +
  M^i_{\mathbf{ts},1}\cdot t_{i,a} +
  M^i_{\mathbf{ts},2}\cdot t_{i,b} +
  M^i_{\mathbf{ts},3}\cdot t_{i,c}
  \cr
s_{i+1,j} &= \tanh(
  M^i_{\mathbf s}\cdot s_{i,j} +
  M^i_{\mathbf{ts}}\cdot\redd_{(a,b,c,g)\in
  F^j_{\mathbf{st}}}(g\cdot x_{i}^{a,b,c})
)\cr
y_{i,d}^{a,b,c,g} &=
  B^i_{\mathbf{st}} +
  M^{1,d}_{\mathbf{st},i}\cdot t_{i,a} +
  M^{2,d}_{\mathbf{st},i}\cdot t_{i,b} +
  M^{3,d}_{\mathbf{st},i}\cdot s_{i,c}\cdot g
  \cr
z_{i,j,d} &= M^i_{\mathbf{st,d}}\cdot \kern -3pt
  \red_{(a,b,c,g)\in F^j_{\mathbf{ts},d}}(
  \ReLU(y_{i,d}^{a,b,c,g})
)\cr
v_{i,j} &= M^i_{\mathbf{tc}}\cdot \red_{a\in F^j_{\mathbf{tc}}}(c_{i,a})\cr
t_{i+1,j} &= \ReLU(
  B^i_{\mathbf t} +
  M^i_{\mathbf t}\cdot t_{i,j} + v_{i,j} + \kern -3pt
  \sum_{d\in\{1,2,3\}} z_{i,j,d}
)\cr
\end{align*}

Here, all the $B$ symbols represent learnable vectors (biases), and all the
$M$ symbols represent learnable matrices. Their sizes are listed in Fig.~\ref{matrice-sizes}.

\begin{figure}
\hbox to\hsize{\hss$
B^i_{\mathbf c} : d^{i+1}_{\mathbf c}
$\hss$
B^i_{\mathbf{ts}} : d^{i+1}_{\mathbf s}
$\hss$
B^i_{\mathbf{st}} : d^{i+1}_{\mathbf t}
$\hss$
B^i_{\mathbf t} : d^{i+1}_{\mathbf t}
$\hss}
\hbox to\hsize{\hss
\vtop{\halign{\hss$#{}$&${}#$\hss\cr
M^i_{\mathbf c}         &: d^{i+1}_{\mathbf c}\times d^i_{\mathbf c}\cr
M^i_{\mathbf{ct}}       &: d^{i+1}_{\mathbf c}\times 2d^i_{\mathbf t}\cr
M^i_{\mathbf s}         &: d^{i+1}_{\mathbf s}\times d^i_{\mathbf s}\cr
}}\hss\vtop{\halign{\hss$#{}$&${}#$\hss\cr
M^i_{\mathbf t}         &: d^{i+1}_{\mathbf t}\times d^i_{\mathbf t}\cr
M^i_{\mathbf{tc}}       &: d^{i+1}_{\mathbf t}\times 2d^i_{\mathbf c}\cr
M^i_{\mathbf{ts}}       &: d^{i+1}_{\mathbf s}\times 2d^{i+1}_{\mathbf s}\cr
}}\hss\vtop{\halign{\hss$#{}$&${}#$\hss\cr
M^i_{\mathbf{ts},j}     &: d^{i+1}_{\mathbf s}\times d^i_{\mathbf t}\cr
M^i_{\mathbf{st},j}     &: d^{i+1}_{\mathbf t}\times 2d^{i+1}_{\mathbf t}\cr
M^{k,j}_{\mathbf{st},i} &: d^{i+1}_{\mathbf t}\times d^i_{\mathbf t}\cr
}}\hss}
    \centering
    \caption{\label{matrice-sizes}Sizes of learnable biases and matrices for $i=0,\ldots,L-1$ and $j,k\in\{1,2,3\}$.}
\end{figure}


By a reduction
operation $\red\nolimits_{i\in I}(u_i)$, where all $u_i$ are vectors of
the same dimension $n$, we mean the vector of dimension $2n$ obtained
by concatenation of $\max_{i\in I}(u_i)$ and $\avg_{i\in I}(u_i)$. The
maximum and average operation are performed point-wise. We also
use another reduction operation $\redd$ defined in the same way except
taking $\max_{i\in I}(u_i)+\min_{i\in I}(u_i)$ instead of just
maximum. This makes $\redd$ commute with multiplication by $-1$.
If a reduction operation obtains an empty input (the indexing set is
an empty set), the result is the zero vector of the expected size.

We construct sets $F^{j_{\mathbf c}}_{\mathbf{ct}}$ and $F^{j_{\mathbf t}}_{\mathbf{tc}}$
based on $E_{\mathbf{ct}}$, and
$F^{j_{\mathbf s}}_{\mathbf{st}}$ and
$F^{j_{\mathbf t}}_{\mathbf{ts},d}$ based on $E_{\mathbf{st}}$, where
$j_{\mathbf c}=1,\ldots,{n_{\mathbf c}}$, $j_{\mathbf s}=1,\ldots,{n_{\mathbf s}}$,
$j_{\mathbf t}=1,\ldots,{n_{\mathbf t}}$, and $d=1,2,3$. 
Informally, the set $\mathbf F^j_{\mathbf {xy}}$ contains the indices related to type $\mathbf y$ for message passing, given the $j$-th receiving node of type $\mathbf x$.
Formally:
\begin{align*}
F^{j}_{\mathbf{ct}} &= \{a : (C_j, T_a)\in E_{\mathbf{ct}}\}\cr
F^{j}_{\mathbf{tc}} &= \{a : (C_a, T_j)\in E_{\mathbf{ct}}\}\cr
F^{j}_{\mathbf{st}} &= \{(a,b,c,g) : (S_j, T_a, T_b, T_c, g)\in E_{\mathbf{st}}\}\cr
F^{j}_{\mathbf{ts},1} &= \{(a,b,c,g) : (S_c, T_j, T_a, T_b, g)\in E_{\mathbf{st}}\}\cr
F^{j}_{\mathbf{ts},2} &= \{(a,b,c,g) : (S_c, T_a, T_j, T_b, g)\in E_{\mathbf{st}}\}\cr
F^{j}_{\mathbf{ts},3} &= \{(a,b,c,g) : (S_c, T_a, T_b, T_j, g)\in E_{\mathbf{st}}\}\cr
\end{align*}

Since $E_{\mathbf{st}}$ can contain a dummy node $T_0$ on the third and fourth positions, following $F^{j}_{\mathbf{st}}$ or $F^{j}_{\mathbf{ts},d}$ in the message passing layer may lead us to a non-existing vector $t_{i,0}$. In that case, we just take the zero vector of dimension $d^i_{\mathbf t}$.

After $L$ message passing layers, we obtain the embeddings $c_{L,j}$, $s_{L,j}$, $t_{L,j}$ of the clauses $C_j$, symbols $S_j$ and terms and literals $T_j$ respectively.

\section{Invariance Properties}
\label{invariance}

By the design of the network, it is apparent that the output is
invariant under the names of the symbols. Indeed, the names are used only
for determining which symbol nodes and term nodes should be the same
and which should be different.

It is also worth noticing that the network is invariant under
reordering of literals in clauses, and under reordering of clauses.
More precisely, if we reorder the clauses
$C_1, \ldots, C_{n_{\mathbf c}}$, then the values
$c_{i,1}, \ldots, c_{i,n_{\mathbf c}}$ are reordered accordingly, and
the values $s_{i,j}, t_{i,j}$ do not change if they still correspond
to the same symbols and terms (they could be also rearranged in
general). This property is clear from the fact that there is no
ordered processing of the data, and the only way how literals are
attributed to clauses is through graph edges which are also unordered.

Finally, the network is also designed to preserve the symmetry under
negation. More precisely, consider replacing every occurrence of a predicate
symbol $S_x$ by the predicate symbol $\neg S_x$ in every clause $C_x$,
and every literal $T_i$. Then the vectors $c_{i,j}$, $t_{i,j}$ do not
change, the vectors 
$s_{i,j}$ do not change either for all
$j \neq x$, and the vector $s_{i,x}$ is multiplied by $-1$.

We show this by induction on the layer $i$. For layer $0$,
this is apparent since the $S_x$ is a predicate symbol, so
$s_{0,x} = \vec 0 = -s_{0,x}$.
Now, let us assume that the claim is true for a layer $i$. 
We follow the computation of the next layer. The symbol
vectors $s_i$ are not used at all in the computation of $c_{i+1,j}$,
so $c_{i+1,j}$ remains the same. For $s_{i+1,j}$ where $j\neq x$, we
don't use $s_{i,x}$ in the formula, and the signs have not changed in
$F^j_{\mathbf{st}}$. Therefore $s_{i+1,j}$ remains the same.
When computing $t_{i+1,j}$, we multiply every $s_{i,c}$ with the
appropriate sign (denoted $g$ in the formula). Since we have replaced every
occurrence of $S_i$ by $\neg S_i$ and kept the other symbols,
the sign $g$ is multiplied by $-1$ if and only if $c = x$, and
therefore the product does not change.
Finally, when computing $s_{i+1,x}$, we follow the formula below:
$$
s_{i+1,x} = \tanh(
  M^i_{\mathbf s}\cdot s_{i,x} +
  M^i_{\mathbf{ts}}\cdot\redd_{(a,b,c,g)\in
  F^x_{\mathbf{st}}}(g\cdot x_{i}^{a,b,c})
)
$$
where $x_{i}^{a,b,c}$ depends only on values $t_{i,j}$, and therefore
was not changed. We can rewrite the formula as
$$
s_{i+1,x} = -\tanh(
  M^i_{\mathbf s}\cdot (-s_{i,x}) +
  M^i_{\mathbf{ts}}\cdot\redd_{(a,b,c,g)\in
  F^x_{\mathbf{st}}}((-g)\cdot x_{i}^{a,b,c})
)
$$
This is because $\tanh$, addition, matrix multiplication, and the
reduction function $\redd$ are compatible with multiplication by $-1$.
In fact, except $\tanh$ they are all linear, thus compatible with
multiplication by any constant, and $\tanh$ is an odd function.
The second formula for $s_{i+1,x}$ can be also seen as a formula
for minus the value of the
original $s_{i+1,x}$ since $-s_{i,x}$ is the original value of
$s_{i,x}$, and $(-g)$ is the original value of $g$. Therefore
$s_{i+1,x}$ was multiplied by $-1$.

\section{Guiding a Connection Tableaux Prover}
\label{leancop-example}

One of the most important 
uses
of machine learning in theorem
proving is guiding the inferences 
done by
the automating theorem
provers. The first application of our proposed model is to guide the
inferences performed by the \lc prover~\cite{OB03}. This line of work
follows our previous experiments with this prover using the XGBoost system
for guidance~\cite{KaliszykUMO18}. In this section, we first give a
brief description of the
\lc prover, then we explain how we fit our network to the \lc prover, and
finally discuss the interaction between the network and the Monte-Carlo
Tree Search that we use.

The \lc prover attempts to prove a given first-order logic problem by first
transforming its negation to a clausal form and then finding a set of instances of
the input clauses that is unsatisfiable. \lc proves the
unsatisfiability by building a connection tableaux, i.e. a tree,\footnote{In some implementations
this is a rooted forest, as there can be multiple literals in the start clause.}
where
every node contains a literal of the following properties.
\begin{itemize}
  \item The root of the tree is an instance of a given initial clause.
  \item The children of every non-leaf node are an instance of an input clause
    (we call such clauses
    axioms). Moreover, one of the child literals must be complementary
    to the node.
  \item Every leaf node is complementary to an element of its path.
\end{itemize}

\begin{figure}
  \includegraphics[width = \hsize]{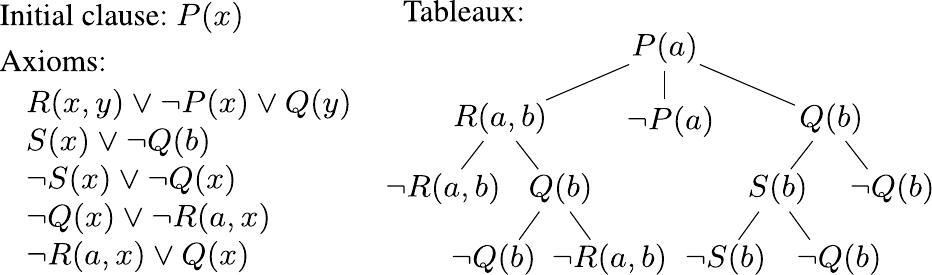}
\caption{Example of a closed connection tableaux, adapted from~\cite{Letz1994ControlledIO}.}\label{tableaux}
\end{figure}

The tree is built during the proof process which involves
automatic computation of substitutions using unification. Therefore the only
decisions that have to be made are ``which axiom should be used for
which node?''. In particular, \lc starts with the initial
clause and in every step, it selects the left-most unclosed (open) leaf. If the
leaf can be unified with an element of the path, the unification is
applied. Otherwise, the leaf has to be unified with a literal in an
axiom, and a decision, which literal in which axiom to use, has to be
made. The instance of the axiom is then added to the tree and the
process continues until the entire tree is closed (i.e., the prover
wins, see Fig. \ref{tableaux}), or there is no remaining available move (i.e., the prover
loses). As most branches are infinite, additional limits are introduced
and the prover also loses if such a limit is reached.
In our experiments, we use a version of the prover with two additional
optimizations: lemmata and regularization, originally proposed by Otten~\cite{DBLP:journals/aicom/Otten10}.

\begin{figure}
  \centerline{\includegraphics[width = 8cm]{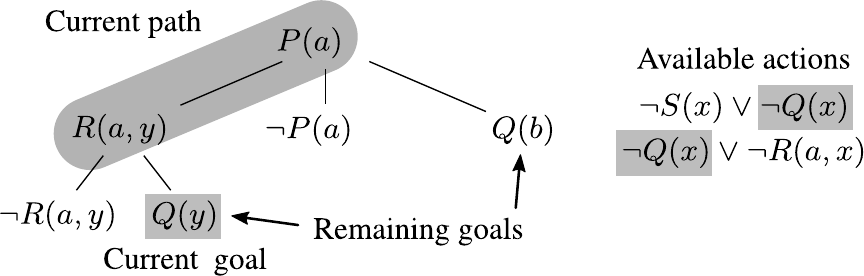}}
  \caption{A state in the \lc solving process}\label{tabstate}
\end{figure}

To guide the proof search (Fig.~\ref{tabstate}), i.e. to select the next action,
we use Monte Carlo Tree Search with policy and value, similar to the AlphaZero~\cite{silver2017mastering} algorithm. This
means that the trainable model should take a \lc state as its
input, and return estimated value of the state, i.e., the probability that the
prover will win, and the action logits, i.e., real numbers assigned to every
available action. The action probabilities are then computed from action
logits using the softmax function.

To process the \lc state with our network, we first need to
convert it to a list of clauses. If there are $A$ axioms, and a
path of length $P$, we give the network the $A+P+1$ clauses: every
axiom is a clause and every element in the path is a clause consisting of
one literal. The last clause given to the network consists of all
the unfinished goals, both under the current path and in earlier
branches. This roughly corresponds to the set of clauses from which we
aim to obtain the contradiction. The initial labels of the clauses can be
therefore of 3 types: a clause originating from a goal, a member of a path, or an
axiom. Each of these types represent a learnable initial vector of
dimension 4.

The symbols can be of two types: predicates and functions, their
initial value is represented by a single real number: zero for
predicates, and a learnable number for functions. For term nodes,
variables in different axioms are always considered to be different,
and they are also considered to be different from the variables in the
tableaux (note that unification performs variable renaming). Variables in the tableaux are shared among the path and the
goals. Every term node can be of four types: a variable in an axiom,
a variable in the tableaux, a literal, or another term. The term nodes have
initial dimension 4.

Afterwards, we propagate through five message passing layers, with
dimensions $d_{\mathbf c}^i = d_{\mathbf c}^i = 32$,
$d_{\mathbf s}^i = 64$, obtaining $c_{5,j}$, $s_{5,j}$ and $t_{5,j}$.
Then we consider all the $c_{5,j}$ vectors, apply
a hidden layer of size 64 with ReLU activation to them, apply
the $\red$ reduction and use one more hidden layer of size 64 with ReLU
activation. The final value 
is then computed by a linear
layer with sigmoid activation.

Given the general setup, we now describe how we compute the logit for an action corresponding to the use of axiom
$C_i$, and complementing its literal $T_j$ with the current goal. Let
$C_k$ represents the clause of all the remaining goals. We concatenate
$c_{5,i}$, $t_{5,j}$ and $c_{5,k}$, process it with a hidden layer
of size 64 with ReLU activation, and then use a linear output layer
(without the activation function).

With the \lc prover, we perform four solving and training
iterations. In every solving iteration, we attempt to solve every problem in
the dataset, generating training data in the meantime. The training
data are then used for training the network, minimizing the
cross-entropy with the target action probabilities and the MSE of the
target value of every trained state. Every solving iteration therefore
produces the target for action policy, and for value estimation, that are used
for the following training.

The solving iteration number 0 (we also call it ``bare prover'') is
performed differently from the 
following ones. We use the prover without guidance, performing random
steps with the step limit 200 repeatedly within a time limit.
For every proof we find, we run the random solver again from every
partial point in the proof, estimating the probabilities that the
particular actions will lead to a solution. This is our training data
for action probabilities. In order to get training data for value, we take all the states which we
estimated during the computation of action probabilities. If the
probability of finding a proof is non-zero in that state, we give it value
1, otherwise, we give it value 0.

Every other solving iteration is based on the network guidance in an
MCTS setting, analogously to AlphaZero~\cite{silver2017mastering} and
to the \rlc system~\cite{KaliszykUMO18}. In order to decide
on the action, we first built a game tree of size 200 according to the PUCT formula
$$
U(s,a) = \log\left(\frac{1+N(s)+c_{\mathrm{base}}}{c_{\mathrm{base}}}\right)\cdot\frac{\sqrt{N(s)}}{1+N(s,a)},
$$
where the prior probabilities and values are given by the network, and
then we select the most visited node (performing a bigstep). This contrasts to the previous
experiment with a simpler clasifier~\cite{KaliszykUMO18} where
every decision node is given 2000 new expansions (in addition to the
expansions already performed on the node).
Additionally a limit of game steps of 200 has been added.
The target probabilities of any state in every bigstep is proportional
the number of visit counts of the appropriate actions in the tree
search. The target value in these states is boolean depending on
whether the proof was ultimately found or not.

\section{DeepMath Experiments}
\label{deepmath-experiments}
DeepMath is a dataset developed for the
first deep learning experiments with premise selection~\cite{IrvingSAECU16ju} on the Mizar40
problems~\cite{KaliszykU13b}. Unlike other datasets such as
HOLStep~\cite{DBLP:conf/iclr/KaliszykCS17}, DeepMath contains first-order formulas
which makes it more suitable for our network. We used the dataset for
two experiments -- \emph{premise selection} (Section~\ref{premise-experiments}) and recovering symbol names from the structure, i.e. \emph{symbol guessing} (Section~\ref{symbol-experiments}).

\subsection{Premise Selection}
\label{premise-experiments}
DeepMath contains 32524 conjectures, and a balanced list of positive
and negative premises for each conjecture. There are on average 8
positive and 8 negative premises for each conjecture. The task we consider first is to
tell apart the positive and negative premises.

For our purposes, we randomly divided the conjectures into 3252 testing
conjectures and 29272 training conjectures. For every conjecture,
we clausified the negated conjecture together with
all its (negative and positive) premises, and gave them all as input to the
network (as a set of clauses). We kept the hyperparameters 
from the \lc experiment. There are two differences. First, there
are just two types of clause nodes: negated conjectures and premises.
Second, we consider just one type of variable nodes.

To obtain the output, we reduce (using the $\red$ function introduced
 in Section~\ref{architecture}) the clause nodes belonging to the conjecture and we do the same
also for each premise. The the two results are concatenated and pushed through a
hidden layer of size 128 with ReLU activation. Finally,
an output layer (with sigmoid activation) is applied to obtain the
estimated probability of the 
premise being positive (i.e., relevant for the conjecture).

\subsection{Recovering Symbol Names from the Structure}
\label{symbol-experiments}
In addition to the standard premise selection task, our setting is
also suitable for defining and experimenting with a novel interesting
task: \emph{guessing the names of the symbols} from the structure of the formula.
In particular, since the network has no information about the names of the symbols, 
it is interesting to see how much the trained system can correctly
guess the exact names of the function and predicates symbols based just on the problem
structure.

One of the interesting uses 
is for \emph{conjecturing by
  analogies}~\cite{GauthierKU16}, i.e., creating new conjectures by
detecting and following alignments of various mathematical theories
and concepts. Typical examples include the alignment between the
theories of additive and multiplicative groups, complex and real
vector spaces, dual operations such as join and meet in lattices, etc.
The first systems used for alignment detection have been so far
manually engineered~\cite{GauthierK19}, whereas in our setting such
alignment is just a byproduct of the structural learning.

There are two ways how a new unique
symbol can arise during the clausification process.
Either as a Skolem function, or as a new definition (predicate)
that represents parts of the original formulas.
We performed two experiments based on how
such new symbols are handled. We either ignore them, and train the neural
network on the original (labeled) symbols only, or we give to all the
new symbols the common labels \texttt{skolem} and
\texttt{def}. Table~\ref{tab:sym-distr} shows the frequencies of
the five most common symbols in the DeepMath dataset after the
clausification. Note that the newly introduced skolems and definitions
account for almost 40\% of the data.

\begin{table}
\setlength{\tabcolsep}{4pt} 
\begin{tabular}{cccccc}
TPTP name & \texttt{def} & \texttt{skolem} & \texttt{=} & \texttt{m1\_subset\_1} & \texttt{k1\_zfmisc\_1} \\
Mizar name & N/A & N/A & \texttt{=} & \texttt{Element} & \texttt{bool} \\
Frequency & 21.5\% & 17.3\% & 2.0\% & 1.7\% & 1.2\%\cr
\end{tabular}
\caption{\label{tab:sym-distr} The most common symbols in the clausified DeepMath.}
\end{table}

%
%

\section{Experimental Results}
\label{results}
\subsection{Guiding \lc}
We evaluate our neural guided \lc against rlCoP~\cite{KaliszykUMO18}. Note however, that for both systems we
use 200 playouts per MCTS decision so the \rlc results presented here are different from~\cite{KaliszykUMO18}. We start with a set of \lc states with their values and action probabilities coming from the 4595 training 
problems 
solved with the bare random prover. 

After training on this set, the MCTS guided
by our network manages to solve 11978 training (160.7\% more) and 1322 (159.2\% more) testing problems, in total 13300 problems (160.5\% more -- Fig.~\ref{invcop}). This is
in total
49.1\%
more than \rlc guided by XGBoost  which in the same setup and with the same limits
solves 8012 training problems, 908 testing problems, and 8920 problems in total. 
The improvement in the first iteration over XGBoost on the training and testing set is 49.5\% and 45.6\% respectively.

Subsequent iterations are also much better than for rlCoP, with
slower progress already in third iteration (note that rlCoP also loses problems, starting with
6th iteration).
The evaluation ran 100 provers in parallel on multiple CPUs 
communicating with the network running on a GPU. 
Receiving the queries from the prover 
takes
on average 0.1 s, while
the message-passing layers alone take around 0.12 s per batch.  The
current main speed issue turned out to be the communication overhead
between the provers and the network. The average inference step in 100
agents and one network inference took on average 0.57 sec.

\begin{figure}
    \centering

\begin{tabular}{ l | r r r r r }
invariant net guided & bare & iter. 1 & iter. 2 & iter. 3 \cr
\hline
overall  & 5105 & 13300 & 14042 & 14002\cr
training & 4595 & 11978 & 12648 & 12642\cr
testing  &  510 &  1322 &  1394 &  1360\cr
\hline
rlCoP & bare & iter. 1 & iter. 2 & iter. 3 \cr
\hline
overall  & 5105 & 8920 & 10030 & 10959 \cr
training & 4595 & 8012 & 9042 & 9874 \cr
testing  &  510 & 908 & 988 & 1085 \cr 
\end{tabular}
\caption{Comparison of the number of problems solved by leanCoP guided by the invariant-preserving GNN and by XGBoost.}
    \label{invcop}
\end{figure}

\bigskip

\begin{figure}
    \centering
    \includegraphics[width=.9\linewidth]{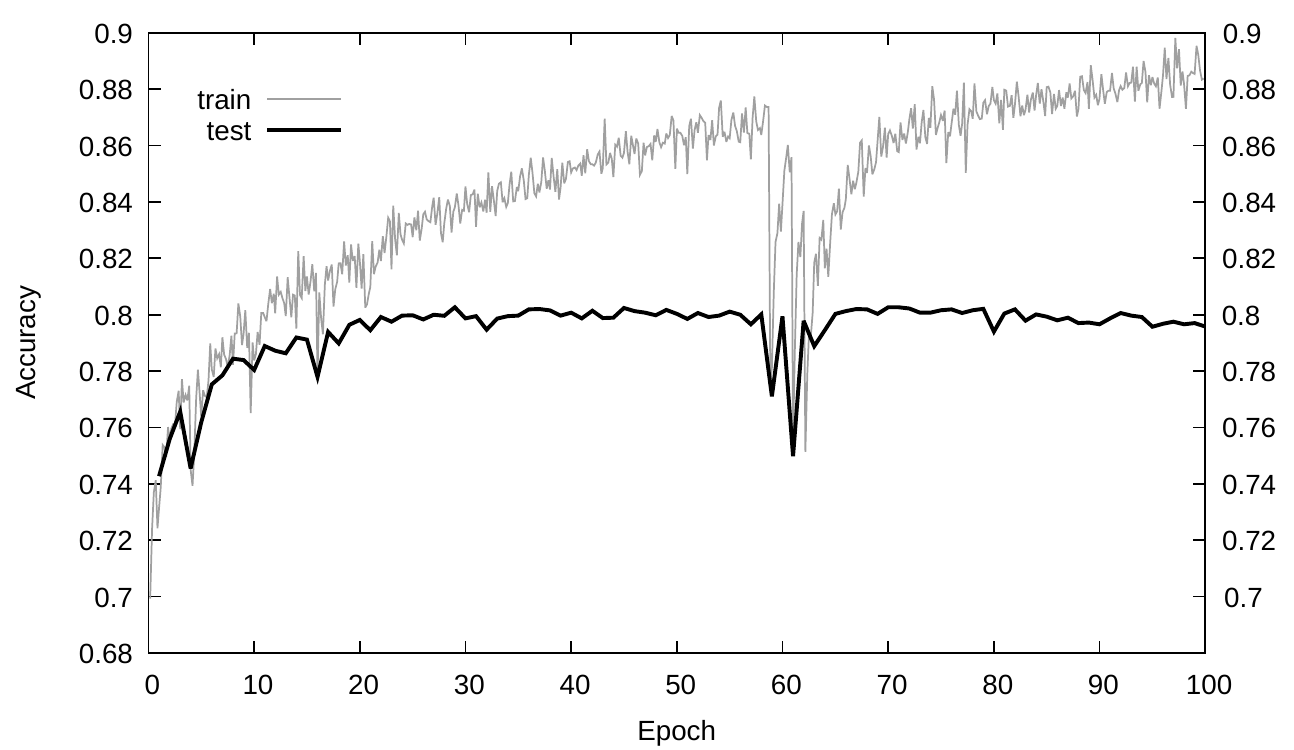}
    \caption{Testing and training accuracy on the premise selection task on the DeepMath dataset.}
    \label{fig:prem_sel}
\end{figure}

\subsection{Premise Selection}
In the first DeepMath experiment with Boolean classification, we obtained testing accuracy of around $80\%$. We trained the network in 100 epochs on minibatches of size 50. A stability issue can be spotted around the epoch 60 from which the network quickly recovered.
We cannot compare the results to the standard methods since the dataset is by design hostile to them -- the negatives samples are based on the KNN, so KNN has accuracy even less than $50\%$. Simpler neural networks were previously tested on the same dataset~\cite{DBLP:journals/corr/abs-1807-10268} 
reaching accuracy $76.45\%$ .

\subsection{Recovering Symbol Names from the Structure}
\begin{figure}
  \centering
    \includegraphics[width=.9\linewidth]{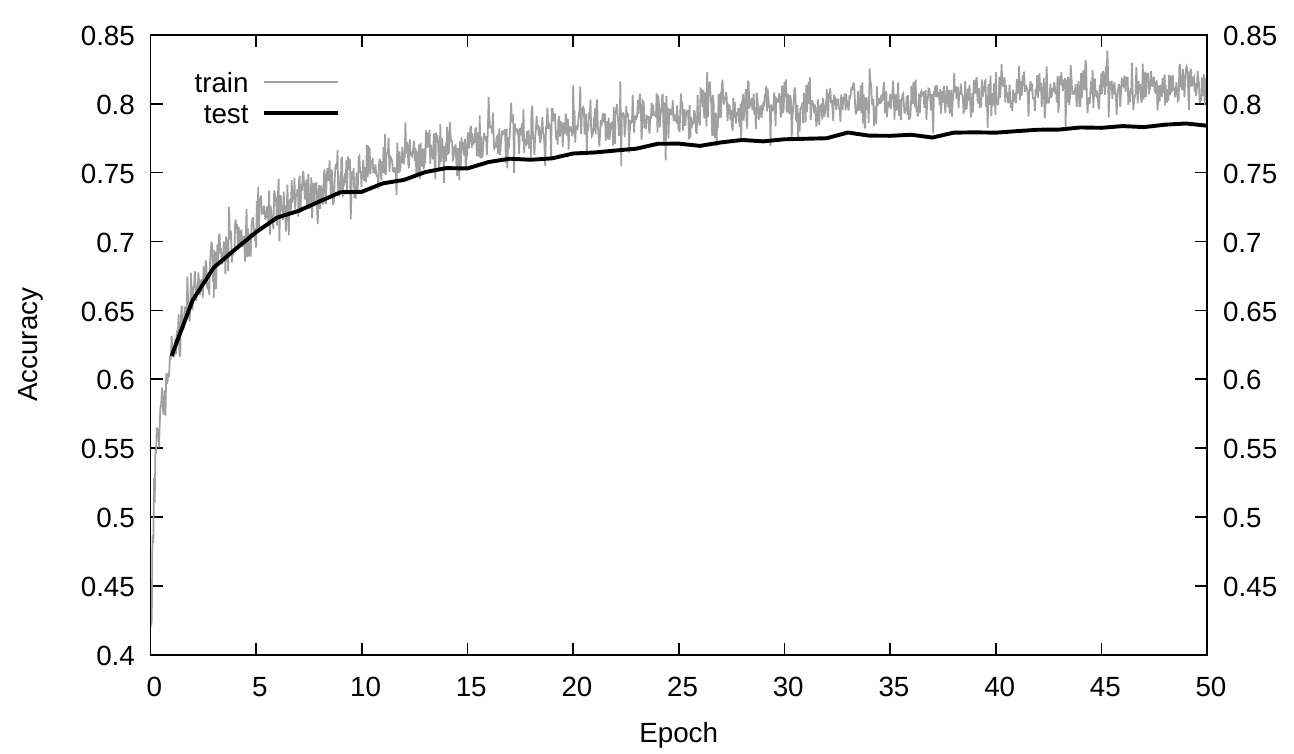}
    \caption{Testing and training accuracy on the label guessing task including labels \texttt{def} and \texttt{skolem} on the DeepMath dataset.}
    \label{fig:symbol_guess_all}
\end{figure}

For guessing of symbol names, we used minibatches consisting only of
10 queries, and trained the network for 50 epochs. When training and
evaluating on the labeled symbols only, the testing accuracy reached
$65.27\%$ in the last epoch. Note that this accuracy is measured on
the whole graph, i.e., we count both the symbols of the conjecture and of the
premises. When training and evaluating also on the
\texttt{def} and \texttt{skolem} symbols, the testing accuracy reached
$78.4\%$ in the last epoch -- see Fig.~\ref{fig:symbol_guess_all}.



\begin{figure}
\centering
  \centering
    \includegraphics[width=.9\linewidth]{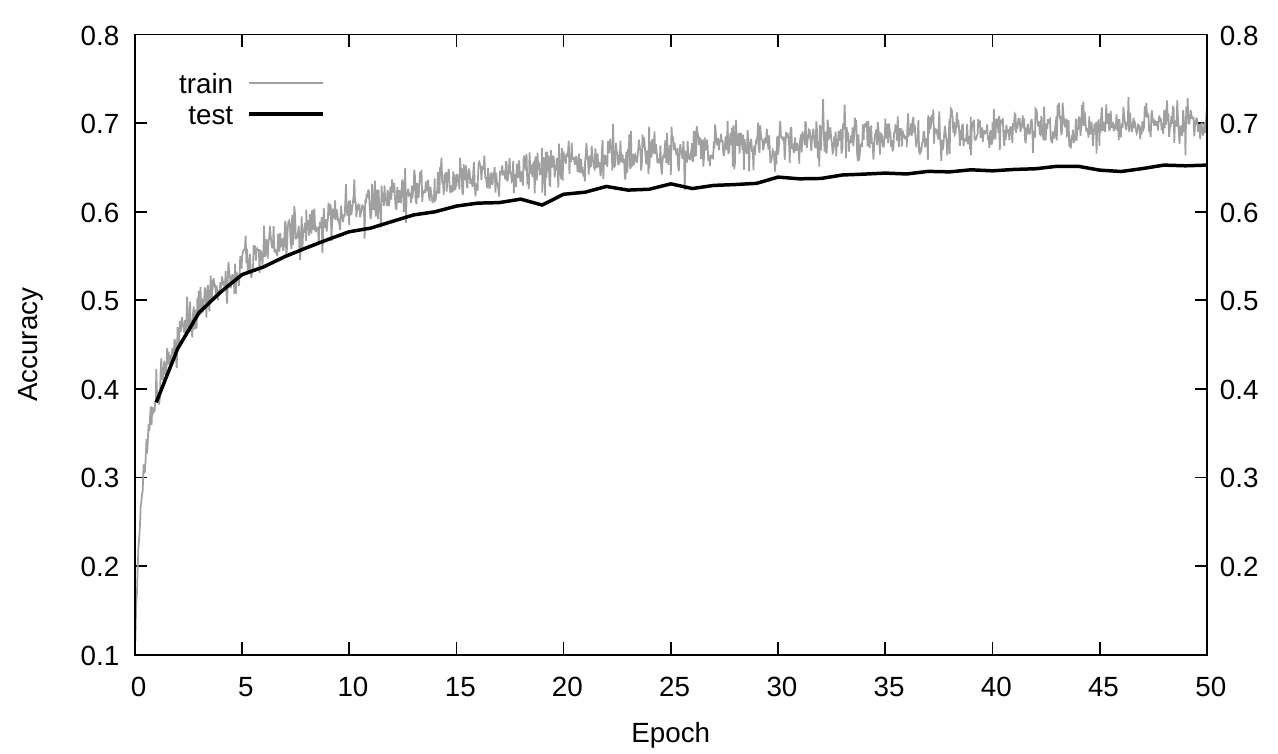}
\caption{Testing and training accuracy on the label guessing task on the DeepMath dataset.}
    \label{fig:symbol_guess_labeled}
  \end{figure}


We evaluate the symbol
guessing (without considering  \texttt{def} and \texttt{skolem}) in more detail on the 3252 test problems and their conjectures.
In particular, for each of these problems and each conjecture symbol,
the evaluation with the trained network gives a list of candidate
symbol names ranked by their probability.  We first compute the number
of cases where the most probable symbol name as suggested by the
trained network is the correct one.  This happens in 22409 cases out
of 32196, i.e., in 70\% cases.\footnote{This differs from the
  testing accuracy of $65.27\%$
  mentioned above, because we only consider the conjecture symbols here.}
A perfect naming of all symbols is achieved for 544 conjectures, i.e.,
in 16.7\% of the test cases. Some of the most common analogies
measured as the common symbol-naming mistakes done on the test
conjectures are shown in Table~\ref{analogies}.
\begin{table}
\begin{center}
\begin{scriptsize}
  \caption{\label{analogies} Some of the common analogies}
\centering
  \begin{tabular}{lllll}
    \toprule
count &   original Mizar symbol &  Mizar analogy \\ 
\midrule
   129& Relation-like&Function-like\\
     69&void&empty\\
     53&Abelian&add-associative\\
     47&total&-defined\\
     45&0&1\\
     40&+&*\\
     39&reflexive&transitive\\
     38&Function-like&FinSequence-like\\
     33&-&+\\
     31&trivial&empty\\
     28& $>=$&=\\
     27&associative&transitive\\
     26&infinite&Function-like\\
     25&empty&degenerated\\
     24&real&natural\\
     23&sigma-multiplicative&compl-closed\\
     20&REAL&COMPLEX\\
     18&transitive&reflexive\\
     18&RelStr&TopStruct\\
     18&Category-like&transitive\\
     17&=&c=\\
     16&initial&infinite\\
     16&[Graph-like]&Function-like\\
     16&associative&Group-like\\
     16&0&\{\}\\
     16&/&*\\
     15&add-associative&associative\\
     15&-&*\\
     13&width&len\\
     13&integer&natural\\
     13&in&c=\\
     12 &$\cap$&$\cup$\\
     11 &c=&$>=$\\
     10 &with\_infima&with\_suprema\\
     10 &ordinal&natural\\
      9 &closed&open\\
      8 &sup&inf\\
      8 &Submodule&Subspace\\
      7 &Int&Cl\\
\bottomrule
  \end{tabular}
\end{scriptsize}
\end{center}
\end{table}

We briefly analyze some of the analogies produced by the network predictions.
In theorem
\texttt{XBOOLE\_1:25}\footnote{\url{http://grid01.ciirc.cvut.cz/~mptp/7.13.01_4.181.1147/html/xboole_1\#T25}}
below, the trained network's best guess correctly labels the symbols
as binary intersection and union (both with probability ca 0.75).
Its second best guess is however also quite probable ($p=0.2$),
swapping the union and intersection. This is quite common, probably
because dual theorems about these two symbols are frequent in the
training data. Interestingly, the second best guess results also in an
provable conjecture, since it easily follows from
\texttt{XBOOLE\_1:25} just by symmetry of equality.
\begin{lstlisting}[language=Mizar,basicstyle=\ttfamily\scriptsize]
theorem :: XBOOLE_1:25
for X, Y, Z being set holds ((X /\ Y) \/ (Y /\ Z)) \/ (Z /\ X) = ((X \/ Y) /\ (Y \/ Z)) /\ (Z \/ X)

second guess:
for X, Y, Z being set holds ((X \/ Y) /\ (Y \/ Z)) /\ (Z \/ X) = ((X /\ Y) \/ (Y /\ Z)) \/ (Z /\ X)
\end{lstlisting}
In theorem
\texttt{CLVECT\_1:72}\footnote{\url{http://grid01.ciirc.cvut.cz/~mptp/7.13.01_4.181.1147/html/clvect_1\#T72}}
the trained network has consistently decided to replace the symbols
defined for complex vector spaces with their analogs defined for real
vector spaces (i.e., those symbols are ranked higher). This is most likely because of the large theory of real
vector spaces in the training data, even though the exact theorem
\texttt{RLSUB\_1:53}\footnote{\url{http://grid01.ciirc.cvut.cz/~mptp/7.13.01_4.181.1147/html/rlsub_1\#T53}} was not among the training data. This again means that the trained network has produced \texttt{RLSUB\_1:53} as a new (provable) conjecture.
\begin{lstlisting}[language=Mizar,basicstyle=\ttfamily\scriptsize]
theorem :: CLVECT_1:72
for V being ComplexLinearSpace for u, v being VECTOR of V
for W being Subspace of V holds 
( u in W iff v + W = (v - u) + W )

theorem :: RLSUB_1:53
for V being RealLinearSpace for u, v being VECTOR of V
for W being Subspace of V holds 
( u in W iff v + W = (v - u) + W )
\end{lstlisting}

Finally, we show below two examples. The first one illustrates on theorems 
\texttt{LATTICE4:15}\footnote{\url{http://grid01.ciirc.cvut.cz/~mptp/7.13.01_4.181.1147/html/lattice4\#T15}} and 
\texttt{LATTICE4:23}\footnote{\url{http://grid01.ciirc.cvut.cz/~mptp/7.13.01_4.181.1147/html/lattice4\#T23}}
the network finding well-known dualities of concepts in lattices (join vs. meet, upper-bounded vs. lower-bounded and related concepts). The second one is an example of a discovered analogy between division and subtraction operations on complex numbers, i.e, conjecturing 
\texttt{MEMBER\_1:130}\footnote{\url{http://grid01.ciirc.cvut.cz/~mptp/7.13.01_4.181.1147/html/member_1\#T130}} from
\texttt{MEMBER\_1:77}\footnote{\url{http://grid01.ciirc.cvut.cz/~mptp/7.13.01_4.181.1147/html/member_1\#T77}}.

\begin{lstlisting}[language=Mizar,basicstyle=\ttfamily\scriptsize]
theorem :: LATTICE4:15
for 0L being lower-bounded Lattice 
for B1, B2 being Finite_Subset of the carrier of 0L holds 
(FinJoin B1) "\/" (FinJoin B2) = FinJoin (B1 \/ B2)

similar to:
theorem Th23: :: LATTICE4:23
for 1L being upper-bounded Lattice
for B1, B2 being Finite_Subset of the carrier of 1L holds 
(FinMeet B1) "/\" (FinMeet B2) = FinMeet (B1 \/ B2)

theorem :: MEMBER_1:77
for a, b, s being complex number holds 
{a,b} -- {s} = {(a - s),(b - s)}

similar to:
theorem :: MEMBER_1:130
for a, b, s being complex number holds 
{a,b} /// {s} = {(a / s),(b / s)}
\end{lstlisting}







\section{Related Work}

Early work on combining machine learning with automated theorem
proving includes,
e.g.,~\cite{DBLP:conf/ogai/ErtelSS89,DenzingerFGS99,DBLP:books/daglib/0002958}.
Machine learning over large formal corpora created from ITP
libraries~\cite{Urban06,MengP08,holyhammer} has been used for premise
selection~\cite{Urb04-MPTP0,US+08-long,KuhlweinLTUH12-long,IrvingSAECU16ju},
resulting in strong \emph{hammer} systems for selecting relevant facts
for proving new conjectures over large formal
libraries~\cite{abs-1108-3446,BlanchetteGKKU16,hh4h4}. More recently,
machine learning has also started to be used to guide the internal
search of the ATP systems. In saturation-style provers this has been
done by feedback loops for strategy
invention~\cite{blistr,JakubuvU18a,SchaferS15} and by using supervised
learning~\cite{JakubuvU17a,LoosISK17} to select the next given
clause~\cite{Overbeek:1974:NCA:321812.321814}. In the simpler
connection tableau systems such as \lc~\cite{OB03} used here, supervised
learning has been used to choose the next tableau extension
step~\cite{UrbanVS11,KaliszykU15}, using Monte-Carlo guided proof
search~\cite{FarberKU17} and reinforcement
learning~\cite{KaliszykUMO18} with fast non-deep learners. Our main evaluation is done in this setting.

Deep neural networks for classification of mathematical formulae were first
introduced in the DeepMath experiments~\cite{IrvingSAECU16ju}
with 1D convolutional networks and LSTM networks.
For higher-order logic, the HolStep~\cite{DBLP:conf/iclr/KaliszykCS17} dataset
was extracted from
interactive theorem prover HOL Light. 1D convolutional neural networks, LSTM, and their combination were proposed as baselines for the dataset.
On this dataset a Graph-based neural network was for a first time applied to theorem proving in the FormulaNet~\cite{DBLP:conf/nips/WangTWD17} work.
FormulaNet, like our work, also represents identical
variables by a single nodes in a graph, being therefore invariant
under variable renaming. Unlike our network, FormulaNet glues
variables only and not more complex terms.
FormulaNet is not designed
specifically for first-order logic, therefore it lacks invariance under
negation and possibly reordering of clauses and literals.
The
greatest difference is however that our network 
abstracts over the symbol names
while FormulaNet learns them individually.

A different invariance property was proposed in a network for propositional calculus by Selsam et
al. \cite{SelsamLBLMD19}. This network is invariant under negation,
order of clauses, and order of literals in clauses, however this 
is restricted to propositional logic, where no quantifiers and variables
are present. 
In the first-order setting, Kucik and
Korovin~\cite{DBLP:journals/corr/abs-1807-10268} performed experiments with basic neural
networks with one hidden layer on the DeepMath dataset.
Neural networks reappeared in state-of-the-art saturation-based proving (E prover) in the work of Loos et al.~\cite{slgicscklpar17}. The considered models included CNNs, LSTMs, dilated convolutions, and tree models.
The first practical comparison of neural networks, XGBoost and Liblinear in guiding E prover was done by Chvalovsky et al. \cite{ChvalovskyJ0U19}.

An alternative to connecting an identifier with all the formulas about it, is to perform definitional embeddings. This has for the first time been done in the context of theorem proving in DeepMath \cite{IrvingSAECU16ju}, however in a non-recursive way. A fully recursive, but non-deep name-independent encoding has been used and evaluated in HOLyHammer experiments~\cite{ckju-mcs-hh}. Similarity between concepts has been discovered using alignments, see e.g.~\cite{DBLP:conf/lpar/GauthierK15}.
Embeddings of particular individual logical concepts have been considered as well, for example polynomials \cite{AllamanisCKS17} or equations \cite{abs-1803-09123}.


\section{Conclusion}

We presented a neural network for processing mathematical formulae invariant under symbol names, negation and ordering of clauses and their literals, and we demonstrated its learning capabilities in
three automated reasoning tasks.
In particular, the network improves over the previous version of \rlc
guided by XGBoost by 45.6\% on the test set in the first iteration of
learning-guided proving. It also outperforms earlier methods on the
premise-selection data, and establishes a strong baseline for symbol
guessing. One of its novel uses proposed here and allowed by this
neural architecture is creating new conjectures by detecting and
following alignments of various mathematical theories and
concepts. This task turns out to be a straightforward application of the structural
learning performed by the network.

Possible future work includes for example 
integration with state-of-the-art saturation-style provers. 
An interesting next step is also evaluation on a heterogeneous dataset such as TPTP where
symbols are not used consistently and learning on multiple libraries --
e.g. jointly on HOL and HOL Light as done previously
by~\cite{DBLP:conf/lpar/GauthierK15} using a hand-crafted alignment
system.

\section{Acknowledgements}
Ol\v{s}\'{a}k and Kaliszyk were supported by the ERC Project \emph{SMART} Starting Grant no. 714034.
Urban was supported by the \textit{AI4REASON} ERC
Consolidator grant number 649043, and by the Czech project
AI\&Reasoning CZ.02.1.01/0.0/0.0/15\_003/0000466 and the European
Regional Development Fund.

\end{document}